# Discriminating between Indo-Aryan Languages Using SVM Ensembles

**Alina Maria Ciobanu[1], Marcos Zampieri[2], Shervin Malmasi[3]**
**Santanu Pal[4], Liviu P. Dinu[1]**

[1]University of Bucharest, Romania, [2]University of Wolverhampton, United Kingdom
[3]Harvard Medical School, United States, [4]Saarland University, Germany
`alina.ciobanu@my.fmi.unibuc.ro`

## Abstract

In this paper we present a system based on SVM ensembles trained on characters and words to discriminate between five similar languages of the Indo-Aryan family: Hindi, Braj Bhasha, Awadhi, Bhojpuri, and Magahi. We investigate the performance of individual features and combine the output of single classifiers to maximize performance. The system competed in the Indo-Aryan Language Identification (ILI) shared task organized within the VarDial Evaluation Campaign 2018. Our best entry in the competition, named ILIdentification, scored 88.95% F1 score and it was ranked 3$^{rd}$ out of 8 teams.

## 1 Introduction

As discussed in a recent survey (Jauhiainen et al., 2018) and in previous work (Tiedemann and Ljubešić, 2012; Goutte et al., 2016), discriminating between similar languages is one of the main challenges in automatic language identification. State-of-the-art n-gram-based language identification systems are able to discriminate between unrelated languages with very high performance but very often struggle to discriminate between similar languages. This challenge motivated the organization of recent evaluation campaigns such as the TweetLID (Zubiaga et al., 2016) which included languages spoken in the Iberian peninsula and the DSL shared tasks (Malmasi et al., 2016b; Zampieri et al., 2015) which included groups of similar languages such as Malay and Indonesian, Bulgarian and Macedonian, and Bosnian, Croatian, and Serbian as well as groups of language varieties such as Brazilian and European Portuguese.

In this paper we revisit the problem of discriminating between similar languages presenting a system to discriminate between five languages of the Indo-Aryan family: Hindi, Braj Bhasha, Awadhi, Bhojpuri, and Magahi. Inspired by systems that performed well in past editions of the DSL shared task such as the one by Malmasi and Dras (2015), we developed a system based on an ensemble of SVM classifiers trained on various groups of word and character features described in more detail in Section 4. Our system competed in the Indo-Aryan Language Identification (ILI) shared task (Zampieri et al., 2018) under the team name ILIdentification. Our best entry achieved performance of 88.95% weighted F1 score and ranked 3$^{rd}$ in the competition.

## 2 Related Work

While some studies focus on increasing the coverage of existing language identification systems by including more languages, as in Brown (2013) and Brown (2014) which include about 1,100 and 1,300 languages respectively, and for the purpose of corpus building (Scannell, 2007), other studies focus on training accurate methods to discriminate between groups of very similar languages such as Indonesian and Malay (Ranaivo-Malançon, 2006), Persian and Dari (Malmasi et al., 2015), and Bosnian, Croatian, Montenegrin, and Serbian (Ljubesic and Kranjcic, 2015).

A first attempting of benchmarking the identification of very similar languages in multilingual settings are the aforementioned Discriminating between Similar Languages (DSL) shared tasks. The DSL tasks have been organized from 2014 (Zampieri et al., 2014) to 2017 (Zampieri et al., 2017b) within the scope of the VarDial workshop series. Four versions of the DSLCC dataset (Tan et al., 2014) have been released containing short excerpts of journalistic texts written in similar languages and language varieties. In the four editions of the DSL shared task a variety of computation methods have been tested. This includes Maximum Entropy (Porta and Sancho, 2014), Prediction by Partial Matching (PPM) (Bobicev, 2015), language model perplexity (Gamallo et al., 2017), SVMs (Purver, 2014), Convolution Neural Networks (CNNs) (Belinkov and Glass, 2016), word-based back-off models (Jauhiainen et al., 2015; Jauhiainen et al., 2016), and classifier ensembles (Malmasi and Dras, 2015), the approach we apply in this paper.

Classifier ensembles showed very good performance not only in language identification but also in similar tasks. Therefore, we build on the experience of our previous work and improve the system that we have previously applied to similar tasks, namely author profiling (Ciobanu et al., 2017) and native language identification (Zampieri et al., 2017a). A detailed description of our system is presented in Section 4.

## 3 Data

The dataset made available by the organizers of the Indo-Aryan Language Identification (ILI) task comprises five similar languages spoken in India: Hindi, Braj Bhasha, Awadhi, Bhojpuri, and Magahi. The process of data collection is described in detail in Kumar et al. (2018). In this paper the authors stress that available language resources are abundant for modern standard Hindi but not for the other four languages included in the dataset. To circumvent this limitation the dataset was compiled primarily by scanning novels, magazines, and newspapers articles using OCR with a subsequent proofreading step in which native speakers proofread the scanned texts to correct OCR mistakes. For Magahi and Bhojpuri texts retrieved from blogs were also included in the dataset.

Over 90,000 documents were made available and the dataset was split into three sets as follows: 70,306 documents were made available for training, 10,329 documents for development, and 9,692 documents for testing. We trained our system using only the data provided by the shared task organizers using no additional training material or external resource.

## 4 Methodology

Following our aforementioned previous work (Ciobanu et al., 2017), we built a classification system based on SVM ensembles using the same methodology proposed by Malmasi and Dras (2015).

The purpose of using classification ensembles is to improve the overall performance and robustness by combining the results of multiple classifiers. Such systems have proved successful not only in NLI and dialect identification, but also in various text classification tasks, such as complex word identification (Malmasi et al., 2016a) and grammatical error diagnosis (Xiang et al., 2015). The classifiers can differ in a wide range of aspects; for example, algorithms, training data, features or parameters.

We implemented our system using the Scikit-learn (Pedregosa et al., 2011) machine learning library, with each classifier in the ensemble using a different type of features. For the individual classifiers, we employed the SVM implementation based on the Liblinear library (Fan et al., 2008), LinearSVC[1], with a linear kernel. This implementation has the advantage of scaling well to large number of samples. For the ensemble, we employed the majority rule VotingClassifier[2], which chooses the label that is predicted by the majority of the classifiers. In case of ties, the ensemble chooses the label based on the ascending sort order of all labels. The individual classifiers were assigned uniform weights in the ensemble.

---

[1] http://scikit-learn.org/stable/modules/generated/sklearn.svm.LinearSVC.html
[2] http://scikit-learn.org/stable/modules/generated/sklearn.ensemble.VotingClassifier.html

We employed the following features:

- Character $n$-grams, with $n$ in $\{1, ..., 8\}$;

- Word $n$-grams, with $n$ in $\{1, 2, 3\}$;

- Word $k$-skip bigrams, with $k$ in $\{1, 2, 3\}$.

We used TF-IDF weighting for all features. In terms of preprocessing, we experimented with punctuation removal, but this did not improve performance.

We first trained a classifier for each type of feature. The individual performance of each classifier is listed in Table 1. The best performing classifier obtains 0.951 F1 score on the development dataset, using character 4-grams as features. Furthermore, we experimented with various ensembles (using various combinations of features) and performed a grid search to determine the optimal value for the SVM regularization parameter $C$, searching in $\{10^{-3}, ..., 10^3\}$. The optimal $C$ value turned out to be 1, and the optimal feature combination was: character bigrams, character trigrams and character 4-grams. With this ensemble, we obtained 0.953 F1 score on the development dataset.

| Feature | F1 (macro) |
|---|---|
| Character 1-grams | 0.599 |
| Character 2-grams | 0.922 |
| Character 3-grams | 0.950 |
| **Character 4-grams** | **0.951** |
| Character 5-grams | 0.943 |
| Character 6-grams | 0.924 |
| Character 7-grams | 0.897 |
| Character 8-grams | 0.866 |
| Word 1-grams | 0.948 |
| Word 2-grams | 0.876 |
| Word 3-grams | 0.639 |
| Word 1-skip bigrams | 0.901 |
| Word 2-skip bigrams | 0.917 |
| Word 3-skip bigrams | 0.920 |

Table 1: Classification F1 score for individual classifiers on the development dataset.

## 5 Results

In this section we report the results obtained using the test set provided by the organizers. We submitted a single run for the ILI task, using the SVM ensemble that obtained the best performance on the development dataset. Our system was ranked 3[rd], obtaining 0.8894 F1 score on the test set. In Table 2 we report the performance of our system and, for comparison, the performance of a random baseline provided by the organizers. Our system outperforms the baseline significantly.

| System | F1 (macro) |
|---|---|
| Random Baseline | 0.202 |
| **SVM Ensemble** | **0.889** |

Table 2: Results for the ILI task on the test dataset.

In Table 3 we present the results obtained by the eight entries that competed in the ILI shared task in terms of F1 score. Teams are ranked according to their performance taking statistical significance into account. ILIdentification was ranked third in the competition below the two teams that were tied in the second place, taraka_rama and XAC, and the best team, SUKI, which outperformed the other participants by a large margin. Team SUKI competed with a system based on the token-based back-off models mentioned in Section 2 (Jauhiainen et al., 2015; Jauhiainen et al., 2016).

| **Rank** | **Team** | **F1 (macro)** |
|---|---|---|
| 1 | SUKI | 0.958 |
| 2 | taraka_rama | 0.902 |
| 2 | XAC | 0.898 |
| **3** | **ILIdentification** | **0.889** |
| 4 | safina | 0.863 |
| 5 | dkosmajac | 0.847 |
| 5 | we_are_indian | 0.836 |
| 6 | LaMa | 0.819 |

Table 3: ILI shared task closed submission rank.

Finally, to better understand the performance of our system on the test set, in Figure 1, we render the confusion matrix of our system. Out of the five classes, BRA is identified correctly most often, while AWA is at the opposite end, with the lowest number of correctly classified instances. Out of the misclassified instances, most are considered to be AWA.

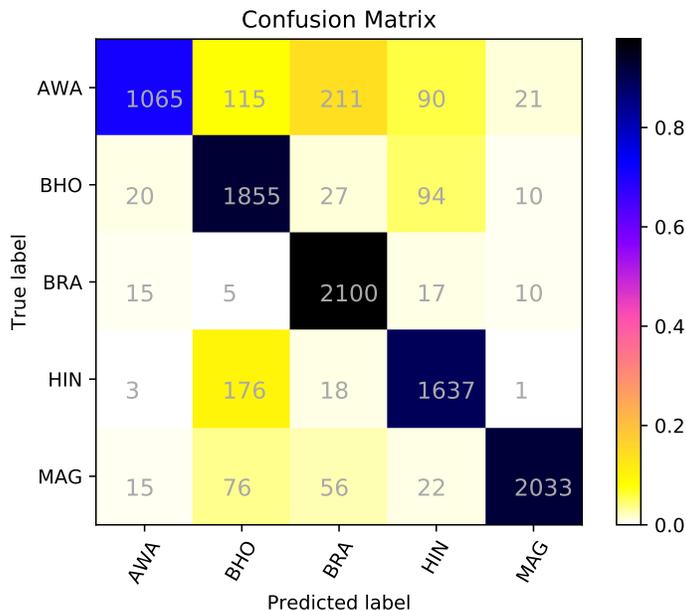

Figure 1: Confusion matrix for the SVM ensemble on ILI task. The languages codes are: Awadhi (AWA), Braj Bhasha (BRA), Bhojpuri (BHO), Hindi (HIN), and Magahi (MAG).

Based on these results, next we present an error analysis of the misclassified instances with the help of two Hindi speakers. This analysis identified a few interesting patterns in system performance and this information can be used to improve the performance of our system.

## 5.1 Error Analysis

In order to better understand the output of the classifiers, we carried out an analysis of the 484 misclassified instances of the development set with the help of two Hindi speakers with some knowledge of the other languages, in particular, of Bhojpuri.

Firstly, we observed that roughly 10% of the misclassified instances were too short containing only one, two or three words. Examples include:

(1) एल । (EN: Have it.)

(2) घर चल । (EN: Lets go home.)

Secondly, the speakers observed that some misclassified examples were named entities such as in the following example:

(3) इण्डियन इंस्टीच्यूट आफ साइन्स एण्ड इंजीनियरिंग न्यू दिल्ली । (EN: Indian Institute of Science and Engineering New Delhi)

Finally, the Hindi speakers observed that Magahi and Bhojpuri instances were very similar, or identical, to Hindi and vice-versa thus making it very challenging for classifiers to discriminate between them. This is particularly true for Bhojpurti, as the confusion matrix presented in Figure 1, shows that 94 Bhojpuri instances were labeled as Hindi and 176 Hindi instances were labeled as Bhojpuri.

One difference observed by the speakers is that Hindi instances often contains a support light verb whereas in Bhojpuri most sentences don't use light verbs. This includes the following example:

(4) सब लोग रामपेआरी के पुतोह समझथ । (EN: Everybody understands Rampyari as their daughter in law.)

## 6 Conclusion and Future Work

This paper presented our submission to the ILI shared task at VarDial 2018. Building on our previous work (Ciobanu et al., 2017), we used an ensemble system consisting of multiple SVM classifiers. Our system ranked third in the competition, obtaining 0.889 F1 score on the test dataset. The features used by the ensemble system were character bigrams, character trigrams, and character 4-grams. Based on the performance on the development dataset reported in Section 4, this was the optimal feature combination.

With the aid of Hindi speakers, in Section 5.1 we presented a concise error analysis of the misclassified instances of the development set. We observed a few interesting patterns in the misclassified instances, most notably that many of the misclassified sentences were too short, containing only one, two or three words, and that several of them contained only named entities. making it very challenging for classifiers to identify the language of these instances.

Another issue discussed in Section 5.1, is that some instances could not be discriminated by native speakers, as noted by Goutte et al. (2016). To cope with these instances one possible direction for future work is to allow a multi-label classification setup in which sentences could be assign to more than one category if annotators labeled them as such.

In future work we would like to explore and compare our methods to other high performance methods for this task. In particular, we would like to try an implementation of the token-based back-off method proposed by the SUKI team. As evidenced in Section 5, SUKI's system achieved substantially higher performance than the other methods in this competition.

## Acknowledgements

We would like to thank the ILI organizers for organizing the shared task and for making the dataset available.